\documentclass[runningheads]{llncs}
\usepackage{graphicx}

\usepackage{tikz}
\usepackage{comment} 
\usepackage{amsmath,amssymb} 
\usepackage{color}
\usepackage{hyperref}

\usepackage{subfig}
\usepackage{textcmds}
\usepackage{multirow}
\usepackage{multicol}
\usepackage{caption}
\usepackage[linesnumbered,ruled,vlined]{algorithm2e}

\begin{document}
\pagestyle{headings}
\mainmatter
\def\ECCVSubNumber{6393}  

\title{Document Structure Extraction using Prior based High Resolution Hierarchical Semantic Segmentation} 

\newcommand*\samethanks[1][\value{footnote}]{\footnotemark[#1]}

\author{Mausoom Sarkar\inst{1} \and
Milan Aggarwal\inst{1}\thanks{equal contribution} \and Arneh Jain\inst{2}\samethanks
 \and Hiresh Gupta\inst{2}\samethanks \and Balaji Krishnamurthy\inst{1}}

\institute{Media and Data Science Research Labs, Adobe \and
Adobe Experience Cloud}

\titlerunning{Hierarchical Document Structure Extraction in High Resolution}
\authorrunning{Sarkar et al.}
%
%
%
\maketitle

\begin{abstract}
Structure extraction from document images has been a long-standing research topic due to its high impact on a wide range of practical applications. In this paper, we share our findings on employing a hierarchical semantic segmentation network for this task of structure extraction. We propose a prior based deep hierarchical CNN network architecture that enables document structure extraction using very high resolution($1800  \times 1000$) images. We divide the document image into overlapping horizontal strips such that the network segments a strip and uses its prediction mask as prior for predicting the segmentation of the subsequent strip. We perform experiments establishing the effectiveness of our strip based network architecture through ablation methods and comparison with low-resolution variations.
Further, to demonstrate our network's capabilities, we train it on only one type of documents (Forms) and achieve state-of-the-art results over other general document datasets. We introduce our new human-annotated forms dataset and show that our method significantly outperforms different segmentation baselines on this dataset in extracting hierarchical structures.Our method is currently being used in Adobe's AEM Forms for automated conversion of paper and PDF forms to modern HTML based forms.
\end{abstract}

\section{Introduction}

Semantic structure extraction for documents has been explored in various works \cite{tablerule,tableprice,historical,colorado}. The task is important for applications such as document retrieval, information extraction, and categorization of content. Document structure extraction is also a key step for digitizing documents to make them reflowable (automatically adapt to different screen sizes), which is useful in web-based services \cite{devices,mobile,dictionary,caseHTML}.  Organizations in domains such as govt services, finance, administration, and healthcare have many documents that they want to digitize. These industries which have been using paper or flat PDF documents would want to re-flow them into digitised version \cite{caseHTML} (such as an HTML) so that they can be used on many devices with different form factors\cite{devices,mobile}. A large part of these documents are forms used to capture data. Forms are complex types of documents because, unlike regular documents, their semantic structure is dense and not dominated by big blobs\footnote{Please refer to supplementary for more visualisations} of structural elements like paragraphs, images.

\begin{figure*}[t]
\centering
\includegraphics[width=\linewidth,height=4cm]{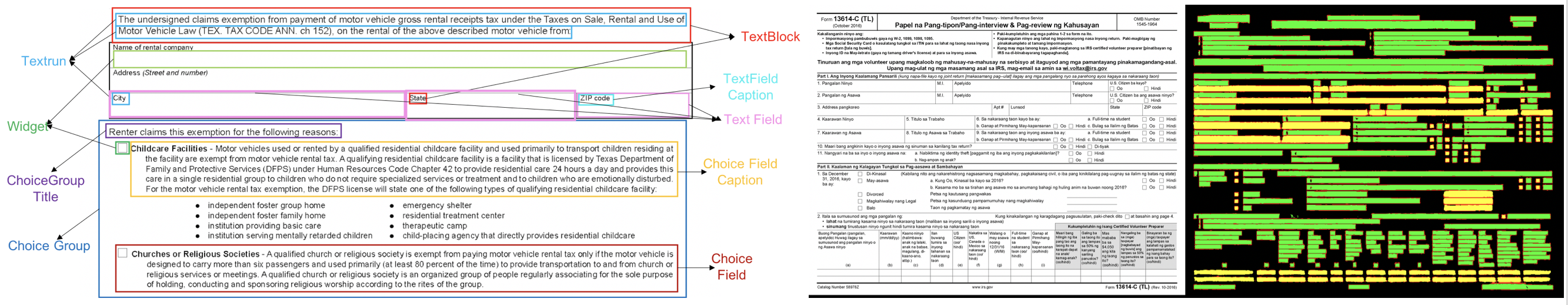}

\caption{(Left): Part of an example form that illustrates elements and structures at different levels of hierarchy. (Right): An illustrative dense form and lower elements segmentation mask predicted by our model. The TextRuns are marked in green and Widgets in yellow.}
\label{fig:describe}
\end{figure*}
\begin{figure*}[t]
\centering
\subfloat[Fragment of a form having large empty spaces]{%
\includegraphics[width=0.4\linewidth,height=0.7in]{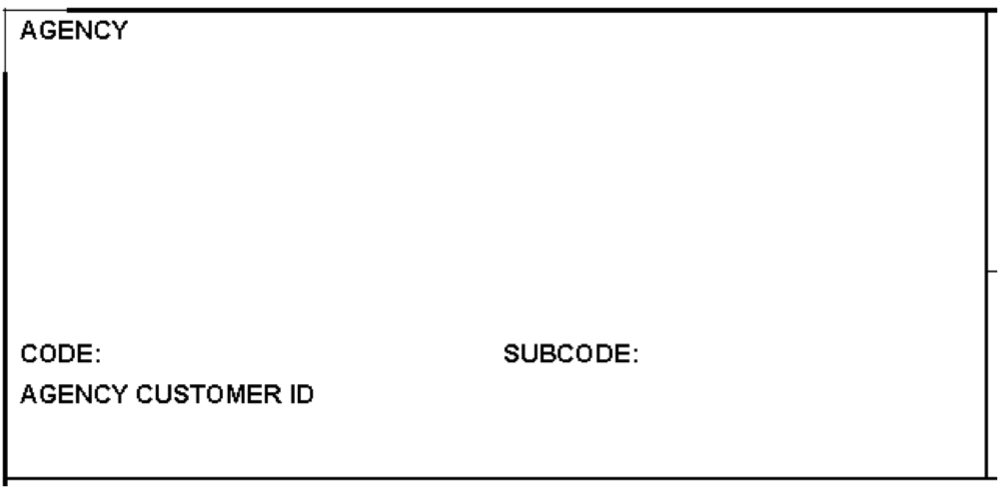}
\label{fig:empty1}
}\hfill
\subfloat[Semantic structure induced around these empty spaces]{%
\includegraphics[width=0.5\linewidth,height=0.75in]{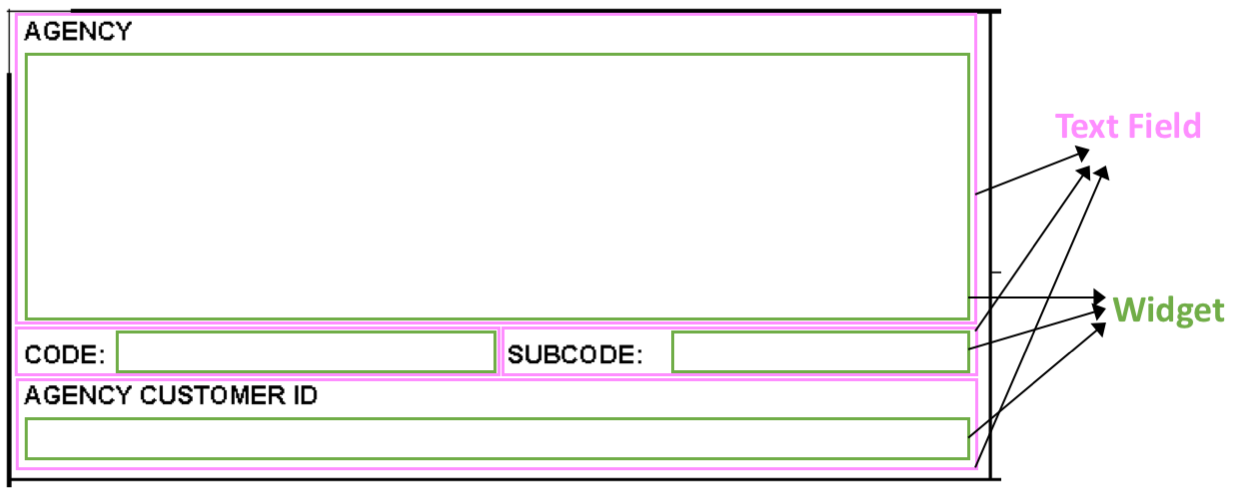}
\label{fig:empty2}
}

\caption{Fragment of a form document}

\label{fig:empty}
\end{figure*}

To make paper or flat-pdf documents reflowable\footnotemark[3], we need to extract its semantic structure at multiple levels of hierarchy. PDFs contain only low-level elementary structures such as text, lines. PDFs do not contain any metadata about other higher-order structures, and therefore, there is a need to retrieve such constructs. Much of previous work looks at regular documents comprising of coarse structural elements that span a large area in the document image, e.g., paragraphs, figures, lists, tables \cite{colorado,schreiber2017deepdesrt,mao2003document}. But, such studies leave out documents having the most complicated structures, i.e., forms. Forms have dense and intricate semantic structures, as shown in Fig \ref{fig:describe}(right). In many forms, the structure is induced due to the presence of large empty areas, like in Fig \ref{fig:empty2}. They also have a deeper hierarchy in structure as compared to other documents, as shown in Fig \ref{fig:describe} (left). We build our method focusing on the hardest cases, i.e., form documents and show that this method generalizes well and establishes new-state-of-art across different document datasets. 

To extract the hierarchical form structure, we identify several composite structures like \textit{TextBlocks}, \textit{Text Fields}, \textit{Choice Fields}, \textit{Choice Groups} that comprise of basic entities like \textit{TextRuns} and \textit{Widgets} as illustrated in Fig \ref{fig:describe} (left). We define a \textit{TextRun} as a group of words present in a single line and \textit{Widgets} as empty spaces provided to fill information in forms. A \textit{TextBlock} is a logical block of self-contained text comprising of one or more \textit{TextRuns}; a \textit{Text Field} comprises of a group of one or more Widgets and a caption \textit{TextBlock} describing the content to be filled in the field. \textit{Choice Fields} are Boolean fields used for acquiring optional information. A \textit{Choice Group} is a collection of such \textit{Choice Fields} and an optional \textit{Choice Group Title}, which is a \textit{TextBlock} that describes instructions regarding filling the \textit{Choice Fields}. Fig \ref{fig:describe} (left) illustrates different semantic structures present in a form document at different hierarchical levels.



We started with fully CNN (FCNNs) based segmentation network since they have been shown to perform well on natural images. However, we found that they perform poorly on form documents as shown in Experiments section. Even FCNNs \cite{tableprice,historical,colorado} focusing on document structure extraction perform well at only extracting coarse structures in documents. They do not perform well at extracting closely spaced structures in form images. Since they process the entire image in a single forward pass, due to memory constraints, they downscale the original image before providing it as input to their model. Moreover, down-scaling of input makes it difficult to disambiguate closely spaced structures, especially in dense regions and leads to merging of different structures. These gaps in current solutions became the motivation behind our current research. In this work, we propose a method to extract the lower-level elements like TextRuns and Widgets along with higher-order structures like Fields, ChoiceGroups, Lists, and Tables. Our key contributions can be summarised as:
\begin{itemize}

\item We propose a prior and sub-strips based segmentation mechanism to train a document segmentation model on very high resolution ($1800\times 1000$) images. Further, our network architecture does not require pre-training with Imagenet \cite{deng2009imagenet} or other large image datasets.
\item We perform hierarchical semantic segmentation and show it leads to better structure extraction for forms. We also compare different variants of our approach to highlight the importance of shared hierarchical features.
\item We propose bi-directional 1d dilated conv blocks to capture axis parallel dependency in documents and show it is better than equivalent 2d dilated conv blocks.
\item We introduce a new human-annotated Forms Image dataset, which contains bounding boxes of a complete hierarchy of semantic classes and all the structures present in the images.
\item We compare our method with semantic segmentation baselines, including DeeplabV3+ (state of the art for semantic segmentation), which uses Imagenet pre-training, outperforming them significantly.
\end{itemize}
\begin{figure}[h]

\centering
\includegraphics[width=\linewidth,height=0.8in]{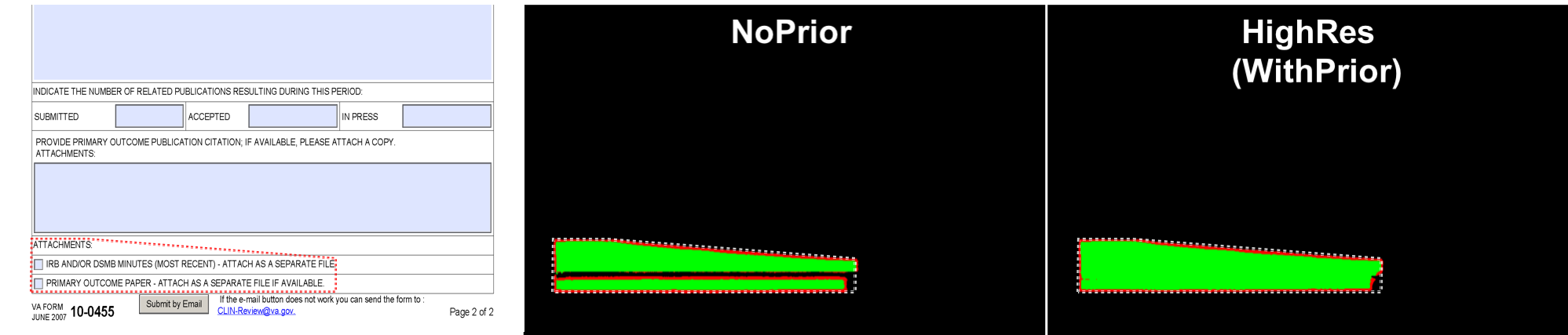}
\caption{HighRes model predicts a single choice group correctly. Two strips cut across group causing NoPrior model to split it since it did not have prior from previous strip. We train our model to predict crisp masks through convex hull.}
\label{fig:noprior}

\end{figure}

Our strip based segmentation helps to mitigate the memory limitation on GPU while training a neural network on high-resolution images. However, strip-based segmentation without prior can potentially fail to predict continuous semantic structures that span across multiple strips (Fig \ref{fig:noprior}). Hence we introduce a prior based strip segmentation, where each image strip's prediction is cached on the GPU and provided as prior concatenated with the input while predicting the segmentation mask of the subsequent strip. Structures that typically span a large area of a form or document like tables and lists could be processed at a lower resolution, but they significantly benefit from the 1D dilated conv network. The hierarchical 1D dilated conv network was introduced to train multi-level hierarchy segmentation together in a single network, so that it learns to predict consistent segmentation masks across these hierarchies \cite{Baxter1997,liu-etal-2015-representation}.

Our method is currently being used in Adobe AEM Forms as Automated Form Conversion Service\footnote{https://docs.adobe.com/content/help/en/aem-forms-automated-conversion-service/using/introduction.html} enabling digitisation to modern HTML based forms.

\section{Related Work}


Document structure analysis started as heuristic-based methods \cite{rule2,rule5,rule3,rule4} based on handcrafted features \cite{rule1} for extracting paragraphs and graphics. Most of the recent deep-learning based approaches are based on fully-convolution neural network (FCN)\cite{colorado,tableprice,historical,chen2014semantic} and avoid any heuristic-based approaches. These FCN's are trained to generate semantic segmentation \cite{FCNSS} for the rasterized version of the document. FCNs have also been used to locate and recognize handwritten annotations in old documents \cite{handwriting}. \cite{curtis} proposed a joint text detection and recognition model. They used a region proposal network which detects the beginnings of text lines, a line following model predicts a sequence of short bounding boxes along the text-line, which is then used to extract and recognize the text. We employ our high resolution segmentation network to extract and disambiguate closely spaced textruns and textblocks from form images.

Table detection has been the key focus of some works like \cite{tablerule,tableprice,tableicpr,tableforback,rastan2019texus}. In \cite{tablerule}, table region candidates were chosen based on some loose rules which were later filtered using a CNN. In \cite{tableprice}, an FCN was proposed, having a multi-scale architecture which had two branches where one was dedicated to table segmentation while the other was used to find contours. After that, an additional CRF(Conditional Random Field) was used to refine the segmentation output further. We propose a multi-branch architecture to segment hierarchical structures that overlap in same region in a form. For tables, we compare our model with \cite{tableprice} on marmot dataset, one of the largest publicly available table evaluation dataset \cite{marmot}. While there are other works \cite{paktable1,schreiber2017deepdesrt} that perform table decomposition into rows and columns (which our model is capable of doing), we discuss table detection only in the scope of this paper. Other works like \cite{figures} introduced a large dataset of 5.5 million document labels focusing on detecting bounding boxes for figures using an Overfeat \cite{overfeat} network, trained over image embedding generated using ResNet-101. 

It is evident that FCN based segmentation approaches have led to great advancement in document structure extraction. However, a few approaches have also tried other network architectures and input modalities such as text. \cite{katti2018chargrid,liu2019graph} are some of the multi-modal approaches proposed to extract named entities from invoices. Other network architectures such as  Graph Neural Networks (GNNs) have been explored in \cite{ribatable,qasim2019rethinking} for detecting tables in invoice documents and parsing table structure, respectively. In a related domain of document classification also, CNN based methods have been explored in recent times. \cite{sicre2017identity} used them for document verification. Moreover, \cite{HAN} proposed HAN to create sentence and document embedding in a hierarchical fashion using a multi-level attention mechanism.
Document classification has also been explored using multi-modal models \cite{docclassmulti} by extracting visual and textual features from MobileNet\cite{mobilenets} and FastText \cite{fasttext} respectively. These features are later on concatenated to learn a better classification model.

In domains such as biomedical imaging and remote sensing, semantic segmentation in high resolution has been explored. \cite{zheng20183} uses past slice’s mask along the z-axis as prior for the entire 2d cross-section and marks out the entire ROI. They convert prior masks into features by a separate net, which are used in decoding. In our approach, each strip prior is partially filled, and only the beginning of ROI is known. Also, we use the mask as prior with the image, which reduces model parameters. Similarly some works use tiles with context mask \cite{januszewski2018high}, but without prior \cite{ronneberger2015u,zhu2017deep}. However for documents, width length slices are required because most context is spread horizontally \& tile-wise context passing would make it harder to understand context spread across the width. \cite{pinheiro2014recurrent,li2016iterative,ren2017end} do iterative refinement of segmentation with different strategies. However, our method does iterative prediction instead of refinement for getting HighRes masks. 


Multi-modal Semantic segmentation has been proposed in \cite{colorado} to extract figures, tables, lists, and several other types of document structures. A text embedding map for the entire page of the document image gets concatenated with the visual feature volume in a spatially coherent manner such that there is a pixel to text correspondence. We use their approach as one of our baselines on forms. We do not compare with object detection methods like \cite{he2017mask} since we found they often merge close structures in dense documents and rather chose better segmentation baseline - DeepLabV3+ \cite{deeplabv3plus2018}, which is the current state-of-the-art in semantic segmentation.





\section{Methodology}
\label{methodology}
In this section, we discuss our proposed model to extract various structures from documents like widgets, fields, textblocks, choice-groups etc. For documents, especially in the case of forms, the semantic structure extends extensively in both vertical and horizontal directions. For many structures such as widgets and fields, it may even extend to empty spaces in the input image requiring the model to predict objects in parts where there is no explicit visual signal. Also, the higher-level structures are composed of lower-level elements and it is necessary to make fine grained predictions at different levels. This leads to our motivation to use a hierarchical dilated 1D conv based semantic segmenter to capture long-range relationships and predict multiple masks at different hierarchies that are mutually consistent. Finally, to address the issue of dense text documents and forms, we modify the network input mechanism by enabling a tile stitching behavior in our network while performing segmentation to train it at higher resolutions.

\subsection{Network Pipeline}
We convert the RGB input image into grayscale and resize the grayscale image having height and width ($I_H \times  I_W$) to ($H \times  w$) such that $I_W$ scales to $w$ and $I_H$ gets scaled by the same ratio, i.e., $w/I_W$. The resulting image is further cropped or padded with zeros to a size of $h \times  w$. $h$ is kept larger than $H$ to accommodate elongated document images. We divide the input image into overlapping horizontal strips. Let $S_h$ be strip height, $O_h$ be overlap height between consecutive strips, $SegNet$ is our segmentation network, and $SegMsk$ denote segmentation mask. Following this notation, Algorithm 1 describes our method where the network predicts the segmentation mask of different strips in succession. Each strip's mask prediction uses the predicted mask for the previous strip as prior. We copy logits corresponding to all classes from segmentation output masks to prior mask having many channels with each channel dedicated to one class.

As stated earlier, we use a dilated 1D conv architecture to predict precise and uniform segmentation masks. Our network broadly comprises of three components, Image Encoder (IE), Context Encoder (CE), and Output Decoder (DE). We concatenate a prior mask to each image strip and feed it into the Image Encoder (IE) to generate features at multiple granular levels. The final features of IE, are then processed through a dilated 1D conv based Context Encoder (CE), which generates features capturing contextual dependencies. All these sets of features from CE and IE is then passed to Output Decoder (DE) to generate segmentation masks for different semantic structure levels. We would now explain each of these modules in greater detail.

\setlength{\textfloatsep}{4pt}
\begin{algorithm}[t]
\caption{Stripwise, prior based image segmentation }
\SetAlgoLined
  \scriptsize 
\textbf{Input}: Image ($Img$) of size $h \times  w$, strip height($S_h$) and overlap height in-between strips($O_h$)\\
\textbf{Output}: Segmentation mask ($OutMsk$) of size $h \times  w$\\
\tcp{$Initialize\ SegMsk,PriorMsk\ and\ StripCount$}
\For{$x\gets0$ \KwTo $w$, $y\gets0$ \KwTo $S_h$}{
$SegMsk[y,x]$ $\leftarrow$ $0$,\\ 
$PriorMsk[y,x]$ $\leftarrow$ $0$,\\
}
$StripCount$ $\leftarrow$ $1+(h-S_h)/(S_h-O_h)$\\
\tcp{$Now\ process\ strips\ one\ by\ one$}
\For{$step\gets0$ \KwTo $StripCount$}{
    \tcp{$Concat\ SegMsk\ with\ PriorMsk$}
    \For{$x\gets0$ \KwTo $w$, $y\gets0$ \KwTo $S_h$}{
    $InpImg[y,x]$ $\leftarrow$ $Img[(S_h-O_h)*step+y,x]\ ||\ PriorMsk[y,x]$
    }
    \tcp{$Predict\ the\ segmentation\ mask\ and \ propagate\ the\ gradients$}
    $SegMsk$ $\leftarrow$ $SegNet(InpImg)$ \label{op0}\\
    \tcp{$Copy\ overlapping\ area\ of\ the\ segmentation \ mask\ into\ prior\ mask$} 
    \For{$x\gets0$ \KwTo $w$, $y\gets0$ \KwTo $O_h$}{
    $PriorMsk[y,x]$ $\leftarrow$ $SegMsk[S_h-O_h+y,x]$
    }
    \tcp{$Calculate\ the\ vertical\ offset\ for\ output$}
    $y\_start$ $\leftarrow$ $(S_h-O_h)*step$\\
    $v_h$ $\leftarrow$ $S_h$\\
    \If{$step<StripCount-1$}{
    $v_h$ $\leftarrow$ $S_h-O_h$
    }
    \tcp{$Collect\ prediction\ in\ OutMsk$}
    \For{$x\gets0$ \KwTo $w$, $y\gets0$ \KwTo $v_h$}{
    $OutMsk[y\_start+y,x]$ $\leftarrow$ $SegMsk[y,x]$
    }
}
 \label{algo:highres_input}

\end{algorithm}
\subsection{Network Architecture}
\begin{figure*}[t]
\centering
\includegraphics[width=\linewidth]{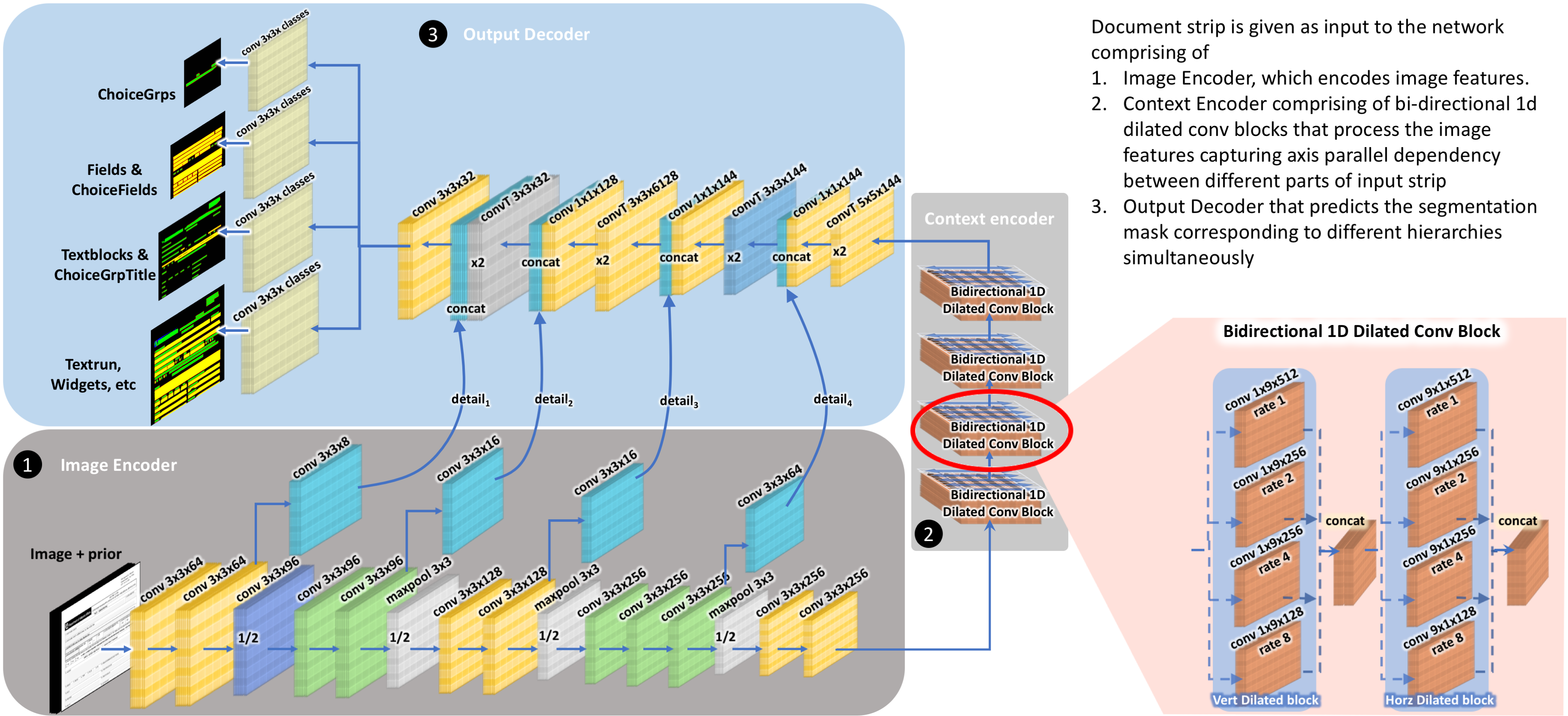}
\caption{Detailed overview of our network architecture.}
\label{fig:network}
\end{figure*}
\subsubsection{Image Encoder}
Fig \ref{fig:network} depicts the architecture of Image Encoder (IE) that comprises of multiple convolution layers, max-pooling layers. As shown in the figure, the first conv layer has $3 \times 3$ kernel with a 64 channel output. The parameters of the remaining layers are highlighted using the same notation. Each convolution layer has a stride of 1 unless specified, the third conv layer in Image Encoder has a stride of 2 and is denoted by "1/2" in the figure. Similarly, all the max-pooling layers have a stride of 2 by default. The output of these convolutions is passed on to the Context Encoder. We extract several intermediate features ($detail_1,detail_2,detail_3,detail_4$) from the Image Encoder that act as skip connections \cite{mao2016image} and are used by the decoder. 

\noindent \textbf{Context Encoder}
The context encoder (CE) is composed of four bidirectional 1D dilated conv blocks (BDB). Each BDB contains a vertical dilated block followed by a horizontal dilated block. The dilated blocks consist of four dilated conv layers \cite{yu2015multi} that work in parallel on the same feature volume at different dilation rates, as seen in Fig \ref{fig:network}.  The BDB processes the feature volume in vertical direction, and subsequently, its outputs are processed in a horizontal direction. Each BDB's output is fed to the next BDB, and final output is fed to a CNN decoder to predict the segmentation mask at all levels of the hierarchy. 

\noindent \textbf{Output Decoder}
The network consists of a single decoder that has multiple heads for generating segmentation maps for different levels in hierarchy. It up-samples it by passing it through a transposed convolution layer \cite{noh2015learning}. The up-sampled features are subsequently passed through another conv layer. Finally, these features are concatenated with another feature volume $detail_4$, obtained from Image Encoder. The decoder branch repeats the sequence of such operations multiple times, as shown in Fig \ref{fig:network}. Each convolution in the decoder branch has a stride of 1, and each transpose conv, depicted as convT in Fig \ref{fig:network}, has a stride of 2 by default. The different segmentation heads on the penultimate layer of the decoder are used to predict segmentation masks for different spatially overlapping classes like widget and fields. The first segmentation head predicts the lowest level of the semantic structure (TextRun and Widget), and the other segmentation heads output prediction corresponding to higher levels of hierarchy. Such a network design helps in segregating the classes according to hierarchy since the container groups, and their constituent classes are predicted in separate masks.

\section{Experiments} \label{Experiments}
\subsection{Datasets}
\noindent \textit{\textbf{Forms Dataset:}}
We used our rich hierarchical Forms Dataset\footnote{A part of the dataset will be made available at https://github.com/flamingo-eccv/flamingo-data} comprising of 52,490 human annotated Form images. These forms are from diverse domains such as automobile, insurance, finance, medical, government (court, military, administration). We employed annotators to annotate the form images to mark the bounding box of different structures in the form image and also asked them to mark the constituent elements that comes lower in the hierarchy for each structure. We split the dataset into 48,256 images for training and 3,234 images for validation. We used a separate set of 1,300 test images for the final evaluation of our model with the baselines and to perform ablation studies. \\
\noindent \textit{\textbf{Marmot Dataset:}}
We evaluate and compare our model trained on Forms Dataset on the Marmot Dataset \cite{marmot}. This dataset is one of the largest publicly available Table evaluation dataset. It contains 2000 document images corresponding to an approximately equal number of English and Chinese documents. \\
\noindent \textit{\textbf{RVL-CDIP:}} The dataset comprises of 400k greyscale images divided into 16 different classes. We select 518 images(mostly scanned) from invoice class annotated with table regions as done by \cite{ribatable} to evaluate and compare our framework on table detection.\\
\noindent \textit{\textbf{ICDAR 2013:}} We also evaluate our approach on the table decomposition task on ICDAR 2013 dataset\cite{gobel2013icdar} where the goal is to decompose tabular regions into rows and columns. It comprises of two sets of pdfs - US and EU split. We extract the images from the pdfs with the corresponding ground truth for tables. We evaluate our model trained a) only on ICDAR datset and b) additional forms data, outperforming state of the art in both settings.

\subsection{Implementation Details}

We set $w=1000$, $h=1800$, $S_h=600$, $O_h=200$ for the $SegNet$ model defined in Section \ref{methodology}. We slice the high resolution input image into $4$ overlapping horizontal strips. All the convolution and deconvolution layers have ReLU activation. We train our model at a batch size of 32 on 8 Tesla V100 GPUs in parallel. We use AdaDeltaOptimizer \cite{zeiler2012adadelta} to train the parameters of our model with an exponentially decaying learning rate using $1 \times 10^{-1}$ as the starting learning rate and a decay factor of $0.1$. Please refer to Fig \ref{fig:network} for specific configuration details of different network layers. To enable the network to predict concise masks, we use convex hull \cite{graham1983finding} to determine segmentation masks.

\subsection{Results}

\noindent \textbf{Model Evaluation and Ablation Studies} \\ 
\noindent On Forms dataset, we train our high resolution model for predicting TextRuns, Widgets, TextBlocks, ChoiceGroup Titles, ChoiceFields, TextFields and Choice Groups such that its decoder predicts structures at various levels of hierarchy. The first hierarchy comprises of TextRuns and Widgets, the second comprises of TextBlocks and ChoiceGroup Titles, the third hierarchy comprises of TextFields and ChoiceFields while the fourth comprises of Choice Groups only. We add another class - Border, surrounding each structure and make the network predict this class to enable it to disambiguate different objects and generalise better.

\noindent We refer to this network configuration as \emph{Highresnet}. We perform ablations establishing gains from our high resolution segmentation network by comparing it with : 1) \emph{Lowresnet} - a low resolution variation of \emph{Highresnet} that takes input image at $792$ resolution and predicts hierarchical segmentation masks for the entire image in a single forward pass; 2) \emph{NoPriorNet} - A \emph{Highresnet} variation where we divide the input image into horizontal strips with no overlap between consecutive strips. In this variant, the segmentation mask predicted for a strip is not given as prior for the subsequent strip prediction; 3) \emph{2D-DilatedNet} - where the horizontal(vertical) 1d dilated conv layers in our network's context encoder are replaced with 2d dilated conv layers with exactly same dilation rates and kernel parameters (each $1\times9$ or $9\times1$ kernel is replaced with $3\times3$ kernel). We use pixel mean Intersection over Union (MIoU) to evaluate different models. We summarise MIoU of different ablations in Table \ref{ablations}. We also estimate object level recall and precision (Table 2) and compare with ablation methods. For this, we consider a predicted structure as correct match if the IoU of its predicted mask is above a threshold (0.7) with an expected structure mask. \textbf{Object-level extraction plays a crucial role in deciding the quality of final re-flow conversion. We, therefore, report these numbers to assess the performance of final structure extraction.}\\
\begin{table*}[t]
 \centering
\caption{Mean IoU of different ablation methods for several hierarchical form structures.}
 \begin{tabular}{cccccccccccc}
 \hline 
  Structure $\rightarrow$ & Text & Widget & Text & ChoiceGroup & Text & Choice & Choice \\
 Model $\downarrow$ & Run & & Block & Title & Field & Field & Group \\
  \hline
  Lowresnet (ours) & 89.31 & 82.17 & 88.49 & 69.03 & 81.93 & 65.85 & 72.61
  \\
  NoPriorNet (ours) & 91.46 & 84.79 & 89.88 & 78.89 & 86.19 & 79.42 & 80.14
  \\
  2D-DilatedNet (ours) & 91.63 & 85.91 & 89.71 & 79.1  & 87.34 & 81.95 & 81.11  
  \\
  Highresnet (ours) & \textbf{92.7} & \textbf{87.32} & \textbf{90.55} & \textbf{80.87} & \textbf{88.87} & \textbf{84.05} & \textbf{83.01}
 \\
  \hline
\end{tabular}

\label{ablations}
\end{table*}
\begin{table*}[t]
\caption{Precision-Recall numbers for the different hierarchical form structures on the different ablation models computed with an IoU threshold of $0.7$.}
\centering
\begin{tabular}{c|c|c|c|c|c|c|c|c|c|c|c|c}
\hline
Model $\rightarrow$ & \multicolumn{3}{c|}{Lowresnet}   & \multicolumn{3}{c|}{NoPriorNet}    & \multicolumn{3}{c|}{2D-DilatedNet} & \multicolumn{3}{c}{Highresnet} \\
 \hline 
      Structure $\downarrow$            & P             & R      & F1       & P             & R    &F1         & P             & R             & F1                 & P                & R             & F1              \\ \hline
TextRun     &72.8&55.0&62.6&79.1&66.9&72.5&80.0&66.7&72.7&\textbf{80.2}&\textbf{67.3}&\textbf{73.2}       \\
Widget       &52.8&51.8&52.3&69.2&70.6&69.9&71.0&71.5&71.2&\textbf{75.0}&\textbf{75.4}&\textbf{75.2}     \\
TextBlock      &51.0&45.6&48.2&69.6&71.6&70.6&68.6&70.4&69.5&\textbf{71.2}&\textbf{72.5}&\textbf{71.9}     \\
Text Field   &43.1&53.4&47.7&66.7&78.0&71.9&69.9&79.7&74.5&\textbf{73.4}&\textbf{82.5}&\textbf{77.7}   \\
ChoiceGroup Title  &48.2&41.0&44.3&83.2&81.8&82.5&82.0&80.8&81.4&\textbf{85.0}&\textbf{84.9}&\textbf{84.9}   \\
Choice Field   &28.3&33.3&30.6&69.8&74.9&72.2&71.7&76.6&74.1&\textbf{77.7}&\textbf{81.5}&\textbf{79.6}  \\
ChoiceGroup   &26.5&33.1&29.4&32.5&43.8&37.3&34.4&43.6&38.5&\textbf{37.8}&\textbf{44.5}&\textbf{40.9}    \\
\hline
\end{tabular}

\label{ablation-pr-1}
\end{table*}
\begin{table*}[t]

\caption{Mean IoU comparison between our Lowresnet model and its variant. Lowresnet-1 to Lowresnet-4 are trained specifically for a single hierarchy only while $Lowresnet_{MD}$ comprises of shared encoder but separate decoders for different hierarchies in a single model.}
 \centering
 \begin{tabular}{cccccccccccc}
 \hline 
  Structure $\rightarrow$ & Text & Widget & Text & ChoiceGroup & Text & Choice & Choice \\
  Model $\downarrow$ & Run & & Block & Title & Field & Field & Group \\
  \hline
  

  Lowresnet-1 (ours) & 87.75 & 80.11 & -- & -- & -- & -- & --
  \\
  Lowresnet-2 (ours) & -- & -- & 87.42 & 63.95 & -- & -- & --
  \\
  Lowresnet-3 (ours) & -- & -- & -- & --  & 80.8 & 63.07 & --  
  \\
  Lowresnet-4 (ours) & -- & -- & -- & --  & -- & -- & 70.55  
  \\
  $Lowresnet_{MD}$ (ours) & 86.42 & 78.96 & 86.87 & 65.67  & 79.28 & 62.79 & 69.37  
  \\
  Lowresnet (ours) & \textbf{89.31} & \textbf{82.17} & \textbf{88.49} & \textbf{69.03} & \textbf{81.93} & \textbf{65.85} & \textbf{72.61}
 \\
  \hline
\end{tabular}
\label{hierarchy}

\end{table*}
\noindent \textbf{Compare $\mathbf{Highresnet}$ with $\mathbf{Lowresnet}$}: It can be seen that by extracting hierarchical structure in high resolution, \emph{Highresnet} is able to improve the MIoU scores significantly over all classes. Similar trend is observed for object level performance (Table 2 and Table 3). \\
\noindent \textbf{Compare $\mathbf{Highresnet}$ with $\mathbf{NoPriorNet}$}: Adding predicted segmentation mask as prior while making prediction for subsequent strip in a page improves the MIoU scores.
Further these improvements in MIoU leads to a significant and even better improvement in object extraction performance(table 2 and 3).\\
\noindent \textbf{Compare $\mathbf{Highresnet}$ with $\mathbf{2D-DilatedNet}$}: It can be seen that using 1d dilated convs performs slightly better than 2d dilated convs(having same number of parameters) in terms of MIoU. However such improvements result in profound impact on object level performance.\\

\noindent \textbf{Ablation on importance of detecting hierarchies simultaneously}: \\
\noindent To analyse the importance of segmenting different hierarchical structures together, we consider different variants of our Lowresnet where we train 4 different models, one for each hierarchy separately: Lowresnet-1 for textruns and widgets, Lowresnet-2 for textblocks and choice group title, Lowresnet-3 for text fields and choice fields, and Lowresnet-4 for choice groups. For these variants we scale down the number of filters in each convolution layer by 2 so that the number of parameters in each variant is scaled down by 4. Since there are 4 such variants, together combined they have same number of parameters as our Lowresnet model. As can be seen in table \ref{hierarchy}, the Lowresnet model has significantly better MIoU for all the structures compared with models trained for individual hierarchy levels. 
This shows that predicting hierarchical structures simultaneously results in better hierarchical features that benefit structures across the hierarchies (for instance, choice group title, choice field and choice group are inter-dependent). We further investigate this through training a variant $Lowresnet_{MD}$ which comprises of the same encoder as in Lowresnet but comprises of four different decoders corresponding to each hierarchy level. The encoder features are shared between different decoders in $Lowresnet_{MD}$. Each decoder has same architecture and parameters as in the decoder of Lowresnet. It can be seen in table \ref{hierarchy} that predicting hierarchical structures together at the last layer in Lowresnet is beneficial compared to having separate decoders in $Lowresnet_{MD}$, even though the latter has more number of trainable parameters. This is because the shared hierarchical features upto the last layer helps in predicting different structures better as compared to having independent features for different hierarchies through separate decoders. 

\begin{figure*}[t]
\centering
\includegraphics[width=\linewidth]{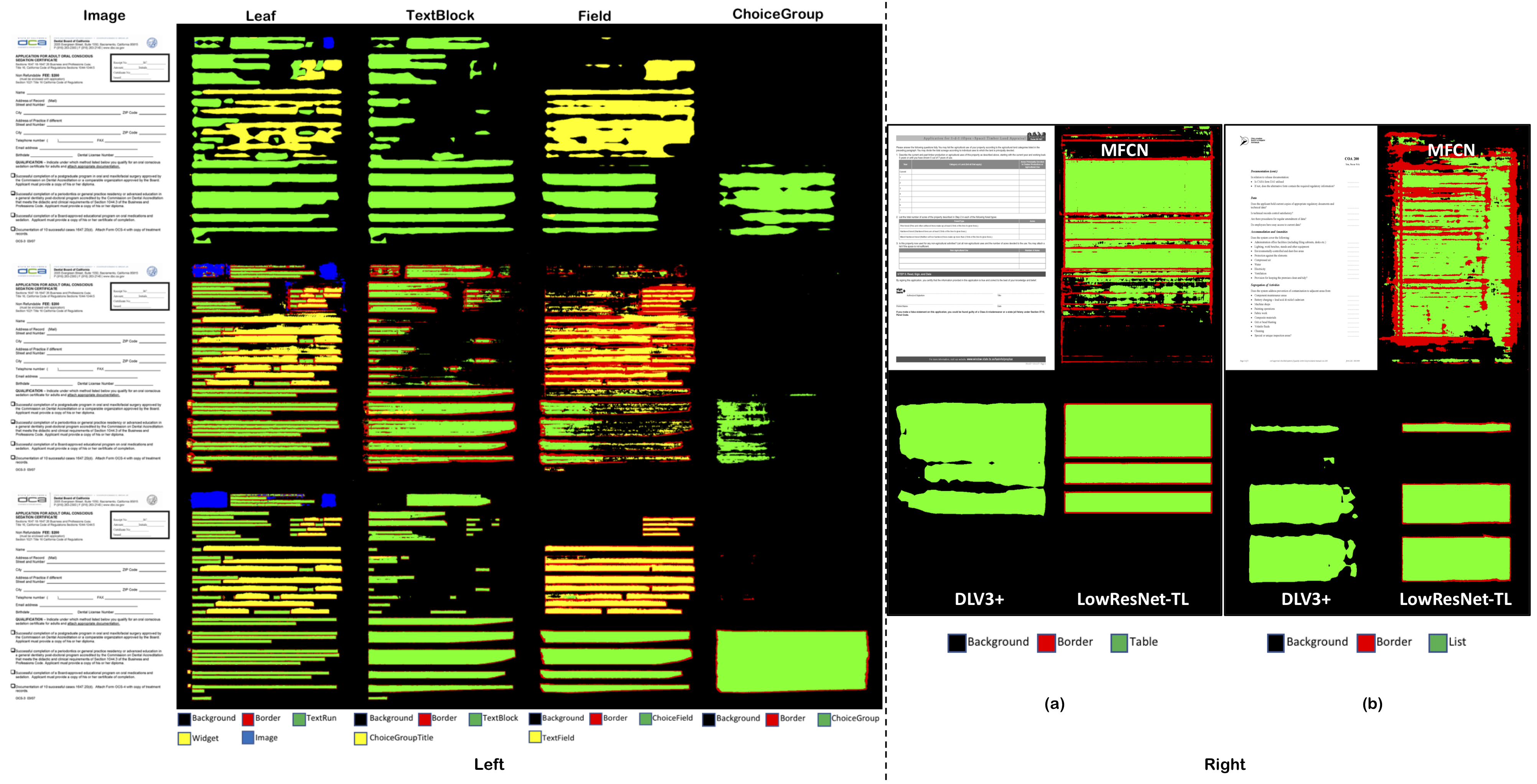}

\caption{Left: Visualisations showing segmentation masks predicted by DLV3+(top row), MFCN(middle row), and our method $Highresnet$(bottom row) for a sample form image. Right: Visualisation of List and Table Segmentation masks on our model and baselines respectively: For each of the two images, MFCN, DLV3+ and Lowresnet-TL predictions are shown in top right, bottom left and bottom right subparts.}
\label{fig:vis1}

\end{figure*}

\noindent \textbf{Comparison with Baselines}
We consider two baselines - DeepLabV3+ \cite{deeplabv3plus2018} (DLV3+), which is the state of the art for semantic segmentation tasks on natural images and Multimodal FCNN (MFCN) \cite{colorado} designed for extracting several complex structures in documents. The baseline segmentation models segment the input image into a flattened hierarchy. To address this, we process output of penultimate layer of the baseline models through 4 separate FC layers to obtain hierarchical masks using data schema similar to \emph{Highresnet}. We train baselines on RGB images at a resolution of $792 \times 792$ following an aspect ratio preserving resize. For DLV3+, we train both with and without imagenet pre-trained weights for the Resnet-101 backbone variants. For MFCN, loss for different classes is scaled according to pixel area covered by elements of each class (calculated over the dataset) as described in their work. Table \ref{baseline} compares the MIoU of our approach with baselines while table \ref{bs-1} compares the object level F1 score. As can be seen, our model \emph{Highresnet} significantly outperforms both baselines on all form structures. In particular, DLV3+ without imagenet pre-training performs poorly on segmenting different form structures. The pre-trained version performs much better but our Highresnet significantly outperforms it without requiring imagenet pre-training with large improvements in MIoU scores.
\begin{table*}[t]

\caption{Mean IoU comparison between the baseline methods and our model on different form structures.}
 \centering
 \begin{tabular}{cccccccccccc}
 \hline 
  Structure $\rightarrow$ & Text & Widget & Text & ChoiceGroup & Text & Choice & Choice \\
 Model $\downarrow$ & Run & & Block & Title & Field & Field & Group \\
  
  \hline 

  DLV3+ NoImagenet & 80.05 & 71.2 & 79.61 & 18.69 & 66.91 & 33.59 & 39.61
  \\
  DLV3+ Imagenet & 81.63 & 77.73 & 83.44 & 48.09 & 76.26 & 50.12 & 56.11
  \\
  MFCN & 77.81 & 47.58 & 71.33 & 29.76  & 39.55 & 28.1 & 35.43  
  \\
  Lowresnet (ours) & 89.31 & 82.17 & 88.49 & 69.03 & 81.93 & 65.85 & 72.61
  \\
  Highresnet (ours) & \textbf{92.7} & \textbf{87.32} & \textbf{90.55} & \textbf{80.87} & \textbf{88.87} & \textbf{84.05} & \textbf{83.01}
 \\
  \hline

\end{tabular}
\label{baseline}

\end{table*}
\begin{table*}[t]
\centering
\caption{Comparison of precision-recall for the different hierarchical form structures between baseline and our method computed with an IoU threshold of $0.7$. *CG Title - ChoiceGroup Title, **CG - ChoiceGroup}
\begin{tabular}{c|c|c|c|c|c|c|c|c|c|c|c|c|c|c|c}
\hline
Model $\rightarrow$ & \multicolumn{3}{|c}{DLV3+}   & \multicolumn{3}{|c}{DLV3+}    & \multicolumn{3}{|c}{MFCN} & \multicolumn{3}{|c}{LowRes} & \multicolumn{3}{|c}{HighRes} \\
 & \multicolumn{3}{|c}{NoImagenet}   & \multicolumn{3}{|c}{Imagenet} & \multicolumn{3}{|c}{} & \multicolumn{3}{|c}{Net (ours)} & \multicolumn{3}{|c}{Net (ours)} \\ 
 \hline 
 Structure $\downarrow$     & P             & R      & F1       & P             & R    &F1         & P             & R             & F1                 & P                & R             & F1  & P                & R             & F1              \\ \hline
TextRun    &57.4&35.1&43.5&63.0&38.1&47.5&58.7&45.9&51.5&72.8&55.0&62.6&\textbf{80.2}&\textbf{67.3}&\textbf{73.2}     \\
Widget       &37.0&32.7&34.7&58.7&53.4&55.9&14.3&26.6&18.6&52.8&51.8&52.3&\textbf{75.0}&\textbf{75.4}&\textbf{75.2}    \\
TextBlock  &52.3&47.5&49.8&57.3&53.3&55.2&17.4&22.3&19.5&51.0&45.6&48.2&\textbf{71.2}&\textbf{72.5}&\textbf{71.9}  \\
TextField   &15.6&14.5&15.0&29.6&23.8&26.4&4.2&27.6&7.4&43.1&53.4&47.7&\textbf{73.4}&\textbf{82.5}&\textbf{77.7}   \\
CG Title*  &46.3&11.2&18.1&59.9&41.3&48.9&10.6&20.8&14.1&48.2&41.0&44.3&\textbf{85.0}&\textbf{84.9}&\textbf{84.9}  \\
ChoiceField   &22.0&14.8&17.7&31.8&23.5&27.0&8.9&19.5&12.2&28.3&33.3&30.6&\textbf{77.7}&\textbf{81.5}&\textbf{79.6} \\
CG**  &4.5&6.8&5.4&14.2&19.4&16.4&1.3&6.0&2.2&26.5&33.1&29.4&\textbf{37.8}&\textbf{44.5}&\textbf{40.9}  \\
\hline
\end{tabular}
\label{bs-1}

\end{table*}

Fig \ref{fig:vis1}(Left) illustrates segmentation masks predicted by different baseline methods, and our model on a sample form image\footnote{Please refer to supplementary for more visualisations}. Baseline methods merge different elements and hierarchical structures such as TextBlocks and Fields. In contrast, our model predicts crisp segmentation masks while extracting all such structures. For choice group, the baseline methods predict incomplete segmentation mask while our model captures long-range dependencies among its constituent elements and predict the complete mask. \\


\begin{table}[t]

\caption{Comparison of MIoU, object level precision, recall, F1 scores of our method with the baselines for Table and List on Forms Dataset.}
 \centering
 \begin{tabular}{c|c|c|c|c|c|c|c|c}
\hline
Model $\rightarrow$ & \multicolumn{2}{|c}{DLV3+}   & \multicolumn{2}{|c}{DLV3+}  & \multicolumn{2}{|c}{MFCN} & \multicolumn{2}{|c}{Lowresnet-TL} \\
Metric $\downarrow$ & \multicolumn{2}{|c}{NoImagenet}   & \multicolumn{2}{|c}{Imagenet}  & \multicolumn{2}{|c}{} & \multicolumn{2}{|c}{} \\
 \hline 
    & Table             & List      & Table       & List             & Table    & List         & Table    & List    \\ \hline
MIoU &69.9&55.7&77.9&65.1&48.1&22.1&\textbf{79.83}&\textbf{63.60} \\ 
P &20.6&17.4&35.2&29.8&4.09&1.49&\textbf{55.71}&\textbf{55.55} \\
R &50.0&26.5&60.4&38.4&59.375&23.67&\textbf{77.20}&\textbf{52.29} \\
F1 &29.2&21.0&44.4&33.6&7.66&2.81&\textbf{62.89}&\textbf{53.73} \\
\hline
\end{tabular}
 \label{mergenet_comparison}
\end{table}

\begin{table}[t]

\caption{Comparison of Table Detection precision-recall numbers on Marmot and RVL-CDIP Datasets and Table Decomposition on ICDAR2013 Dataset}
 \centering
 \begin{tabular}{c|c|c|c|c|c|c|c|c|c|c|c|c}
 \hline 
 \multirow{2}{*}{Method (IoU)} & \multicolumn{3}{|c|}{Marmot English} & \multicolumn{3}{|c|}{Marmot Chinese} & \multicolumn{3}{|c|}{RVL-CDIP}& \multicolumn{3}{|c}{ICDAR2013}\\
 \cline{2-13}
 
 & P & R & F1 & P & R &F1 & P &R &F1& P &R &F1\\
 \hline
 \hline
 \emph{MSMT-FCN}(0.8)  & \textbf{75.3} & 70.0 & 72.5 & \textbf{77.0} & 76.1 & \textbf{76.5} & -- & -- & --& -- & -- & --\\ 
 \emph{MSMT-FCN}(0.9)  & 47.0 & 45.0 & 45.9 & 49.3 & 49.1 & 49.1 & -- & -- & --& -- & -- & --\\
 \emph{Lowresnet-TL}(0.8) & 75.2 & \textbf{72.2} & \textbf{73.7} & 71.7 & \textbf{77.4} & 74.4 & -- & -- & --& -- & -- & --\\
 \emph{Lowresnet-TL}(0.9) & \textbf{61.2} & \textbf{64.6} & \textbf{62.8} & \textbf{62.3} & \textbf{70.5} & \textbf{66.1} & -- & -- & --& -- & -- & --\\
 \hline
 \emph{GNN-Net\cite{ribatable}}(0.5) & -- & -- & -- & -- & -- & -- &  25.2 & 39.6 & 30.8 & -- & -- & --\\
  
  \emph{Lowresnet-TL }(0.5) & -- & -- & -- & -- & -- & -- &  \textbf{43.6} & \textbf{65.4} & \textbf{52.3} & -- & -- & --\\
  \hline
 \emph{Baseline \cite{siddiqui2019rethinking}}(0.5) & -- & -- & -- & -- & -- & -- & -- & -- & -- & 93.4 & 93.4 & 93.4 \\
 \emph{Lowresnet-TL}(0.5) & -- & -- & -- & -- & -- & --  & -- & -- & -- & \textbf{93.9} & \textbf{94.3} & \textbf{94.1}\\
 \emph{+FormsData}(0.5) & -- & -- & -- & -- & -- & -- & -- & -- & --& \textbf{94.7} & \textbf{95.7} & \textbf{95.2} \\
  \hline
 \end{tabular}
 
 \label{marmot}
\end{table}


\noindent \textbf{Evaluation on Other Higher Order Constructs}
In this section, we discuss the performance of our model at extracting other higher order structures like Lists and Tables. These structures are relatively more evident and span large regions in a page reducing the need to disambiguate them in high resolution. Consequently, we train a separate low resolution ( $792 \times 792$ ) version of our proposed network similar to \emph{Lowresnet} which we refer to as \emph{Lowresnet-TL} to predict these structures. In order to compare the performance of this network, we also train networks for the two baselines for extracting Tables and Lists simultaneously.
Table \ref{mergenet_comparison} compares the MIoU of our method with the baseline models and the Fig \ref{fig:vis1} (Right) illustrates the network outputs for the task of Table and List segmentation. It can be seen that \emph{Lowresnet-TL} significantly outperforms both the baselines, specifically it outperforms imagenet pre-trained DLV3+ while itself not requiring imagenet pre-training.

We also compare precision-recall of Table predictions of \emph{Lowresnet-TL} on Marmot Dataset with previous best method -- Multi-Scale Multi-Task FCN (\emph{MSMT-FCN}) \cite{tableprice} in Table \ref{marmot}. It can be seen that \emph{Lowresnet-TL} performs similar to \emph{MSMT-FCN} for an IoU threshold of $0.8$. However, \emph{Lowresnet-TL} performs significantly better than \emph{MSMT-FCN} for higher IoU threshold of $0.9$ indicating our architecture is able to predict crisper predictions. On RVL-CDIP, our model outperforms GNN-Net \cite{ribatable} which is state of the art for table detection on this dataset. We also evaluate our method for task of table decomposition into rows an columns on ICDAR 2013 dataset and compare it against \cite{siddiqui2019rethinking}. The numbers reported are average of precision, recall and F1 obtained for rows and columns as done by \cite{siddiqui2019rethinking}. We train our model in two settings - one using ICDAR2013 data only(using same train-test split) and secondly by adding our forms data (105 tables). We apply post processing on network outputs where we filter row predictions based on area threshold and extend row mask horizontally to obtain completed row predictions and apply similar transformation for columns. As can be seen our method outperforms \cite{siddiqui2019rethinking}(table \ref{marmot}).

\section{Conclusion}

We propose a novel neural network training mechanism to extract document structure on very high resolution. We observe that higher resolution segmentation is beneficial for extracting structure, particularly on forms since they posses highly dense regions. We show that a single network hierarchical segmentation approach leads to better results on structure extraction task. In addition, we also show that 1D dilated conv based model captures long range contextual dependencies while segmenting different hierarchical constructs. Various ablation studies show the effectiveness of our high resolution segmentation approach and network architecture design. We compare our method with different semantic segmentation baselines outperforming them significantly on our Forms Dataset for several structures such as TextBlocks, Fields, Choice Groups etc. Additionally, our model trained on Forms Dataset outperforms prior art for table detection on Marmot and ICDAR 2013 dataset.
%
%
\bibliographystyle{splncs04}
\bibliography{eccv2020submissionCR}
\end{document}